\documentclass[aos,preprint]{imsart}

\RequirePackage{amsthm,amsmath,amsfonts,amssymb,xcolor}
\RequirePackage[authoryear]{natbib}
\RequirePackage[colorlinks,citecolor=blue,urlcolor=blue]{hyperref}
\RequirePackage{graphicx}
\usepackage{subfigure}
\usepackage{float}
\usepackage{multirow}
\usepackage{diagbox}
\usepackage[figuresright]{rotating}
\usepackage{pdflscape}
\usepackage{enumitem}
\usepackage{mathrsfs}
\usepackage{natbib,graphicx,lscape,longtable}
\usepackage{mathrsfs,amsmath,amsthm,amssymb,xcolor}
\usepackage{natbib,epsfig,graphicx,pdfpages}
\usepackage{rotating}
\usepackage{float}
\usepackage{bm}
\usepackage{ulem}
\usepackage{CJK}
\usepackage{natbib}
\usepackage{algorithm}
\usepackage{algorithmic}
\usepackage{booktabs}
\usepackage{bm}
\usepackage{subfigure}
\usepackage{makecell}
\usepackage{hyperref}
\usepackage{multirow}
\usepackage{threeparttable}
\usepackage{xr}
\usepackage{mathtools}
\startlocaldefs
\theoremstyle{plain}

\newtheorem{theorem}{Theorem}[section]
\newtheorem{lemma}{Lemma}[section]

\theoremstyle{remark}

\newtheorem{assumption}{Assumption}
\newtheorem{corollary}{Corollary}[section]
\newtheorem{proposition}{Proposition}[section]

\def\beq{\begin{equation}}
	\def\eeq{\end{equation}}
\def\beqr{\begin{eqnarray}}
	\def\eeqr{\end{eqnarray}}
\def\beqrs{\begin{eqnarray*}}
	\def\eeqrs{\end{eqnarray*}}
\def\bet{\begin{theorem}}
	\def\eet{\end{theorem}}
\def\bel{\begin{lemma}}
	\def\eel{\end{lemma}}
\def\bes{\begin{step}}
	\def\ees{\end{step}}
\def\bep{\begin{proposition}}
	\def\eep{\end{proposition}}
\def\bg{\begin{figure}[tbph]\begin{center}}
		\def\eg{\end{center}\end{figure}}

\def\bc{\begin{center}}
	\def\ec{\end{center}}

\def\wh{\widehat}

\def\mD{\mathcal D}
\def\mW{\mathcal W}
\def\mQ{\mathcal Q}
\def\mR{\mathbb{R}}
\def\mS{\mathcal S}

\def\mG{\mathcal G}

\def\1{\mbox{\boldmath $1$}}

\def\mH{\mathcal H}

\def\mF{\mathcal F}
\def\mI{\mathcal I}

\def\mT{\mathcal T}

\def\var{\mbox{var}}

\def\E{\mbox{E}}

\def\argmin{\mbox{argmin}}

\usepackage{xr-hyper}
\usepackage{hyperref}
\makeatletter
\newcommand*{\addFileDependency}[1]{
\typeout{(#1)}
\@addtofilelist{#1}
\IfFileExists{#1}{}{\typeout{No file #1.}}
}
\makeatother

\newcommand*{\myexternaldocument}[1]{%
\externaldocument{#1}%
\addFileDependency{#1.tex}%
\addFileDependency{#1.aux}%
}

\myexternaldocument{suppmain}

\begin{document}

\begin{frontmatter}
\title{Optimal Deep Nonparametric Regression with Uniformly Bounded Lipschitz Constants}
\runtitle{Deep Regression with Bounded Lipschitz}

\begin{aug}
\author[A]{\fnms{Jiaxin}~\snm{Shi}\ead[label=e1]{jxshi@stu.pku.edu.cn}},
\author[B]{\fnms{Ziyuan}~\snm{Chen}\ead[label=e2]{chenziyuan@pku.edu.cn}},
\author[C]{\fnms{Jian}~\snm{Huang}\ead[label=e3]{j.huang@polyu.edu.hk}},
\and
\author[A]{\fnms{Hansheng}~\snm{Wang}\ead[label=e4]{hansheng@pku.edu.cn}}

	\address[A]{Guanghua School of Management, Peking University, Beijing, China\printead[presep={,\ }]{e1,e4}}
	
	\address[B]{School of Mathematical Sciences, Peking University, Beijing, China\printead[presep={,\ }]{e2}}
	
	\address[C]{Department of Data Science and Artificial Intelligence, Department of Applied Mathematics, \\
	Hong Kong Polytechnic University, Hong Kong SAR, China\printead[presep={,\ }]{e3}}
\end{aug}

\begin{abstract} 

Stability under input perturbations is necessary for deep neural network estimators in many applications involving noisy covariates or estimated features. Weight norm constraints are widely used to control sensitivity to input perturbations by bounding the Lipschitz constant of neural networks. However, existing approximation results under fixed bounds on standard norm products are insufficient to establish optimal statistical rates. In this paper, we establish a sharp approximation theory for deep ReLU networks under a split spectral norm product constraint. This theoretical result shows that sufficiently smooth H\"{o}lder functions can be uniformly approximated to arbitrary accuracy as the network depth increases, with the norm-product bound held fixed. We then propose a penalized least squares estimator for deep nonparametric regression. The resulting estimator has a uniformly bounded Lipschitz constant while still achieving the optimal nonparametric rate up to a logarithmic factor. As applications, we establish optimal prediction rates for two-stage learning and provide robustness guarantees under adversarial input perturbations. Numerical studies show that the penalized estimator has well-controlled sensitivity to input perturbations and prediction accuracy comparable to that of classical deep network methods.

\end{abstract}

\begin{keyword}[class=MSC]
\kwd[Primary ]{62G08}
\kwd[; secondary ]{68T07}
\kwd{41A25}
\end{keyword}

\begin{keyword}
\kwd{Approximation theory}
\kwd{Deep neural networks}
\kwd{Penalized M-estimation}
\kwd{Non-asymptotic error bound}
\end{keyword}

\end{frontmatter}


\section{Introduction}\label{Intro}

Deep neural networks have achieved remarkable empirical performance in a wide range of tasks, such as computer vision \citep{krizhevsky2012imagenet,he2016deep,kirillov2023segment}, natural language processing \citep{devlin2019bert,openaigpt,guo2025deepseek}, and generative modeling \citep{goodfellow2014generative,ho2020denoising,peebles2023scalable}. In nonparametric regression, neural networks provide a flexible class of estimators. A substantial literature has shown that suitably chosen neural network estimators can achieve minimax prediction rates over smooth function classes, up to logarithmic factors \citep{schmidt2020nonparametric,farrell2021deep,jiao2023deep,bhattacharya2024deep}. 

However, the practical use of neural networks usually involves additional considerations beyond prediction rates. One important property is stability under input perturbations, which is commonly quantified by the Lipschitz constant of the estimator \citep{bartlett2017spectrally,jiao2023approximation}. Such stability is particularly relevant in two-stage learning, where neural networks are used for second-stage prediction with first-stage estimates as inputs. For example, large-scale pretraining provides feature representations for downstream neural predictors  \citep{chen2020simple,oquab2024dinov2}. In such settings, errors in the learned representations become input perturbations to the downstream network. Ideally, small perturbations should not lead to large changes in the final output. Related considerations arise in adversarial robustness, where Lipschitz control limits changes in predictions under bounded input perturbations \citep{goodfellow2014explaining,araujo2023unified}. Lipschitz constraints also play an important role in the formulation of Wasserstein-type generative models \citep{arjovsky2017wasserstein,biau2021some}. Therefore, it is very important to control the Lipschitz constants of neural network estimators for accurate prediction in the presence of input perturbations.

Controlling the Lipschitz constant of a neural network is not straightforward. In practice, fitted neural networks can have large Lipschitz constants and amplify small input perturbations \citep{virmaux2018lipschitz,gupta2021adversarial,bubeck2023universal}. Imposing a constraint based on the exact Lipschitz constant is difficult, since computing the Lipschitz constant is NP-hard in general \citep{virmaux2018lipschitz}. For networks with the rectified linear unit (ReLU) activation, products of layer norms provide computable upper bounds on this constant \citep{bartlett2017spectrally,virmaux2018lipschitz}. Therefore, controlling these products becomes a mainstream way to control input sensitivity. A simple and direct approach is to bound the spectral norm of the weight matrix at each layer by one, which makes the whole network $1$-Lipschitz \citep{cisse2017parseval,miyato2018spectral,gouk2021regularisation}. However, these layerwise constraints can restrict the expressive power of ReLU networks, even for simple Lipschitz functions such as the absolute value function \citep{huster2018limitations,anil2019sorting}. This limitation has motivated alternative architectures based on sorting activations \citep{anil2019sorting} and dynamical systems \citep{meunier2022dynamical}. In addition, \cite{liu2022learning} proposed a regularization method based on norm products. These developments motivate theoretical studies of ReLU networks under norm-product constraints. \cite{jiao2023approximation} established approximation bounds under product constraints based on induced $\ell_\infty$ matrix norms. Under a fixed norm budget, their approximation upper bound does not decrease as the network size increases. Following this line, \cite{yang2024nonparametric,yang2025optimal} obtained sharper bounds for shallow norm-constrained networks. However, their approximation analysis still relies on shallow constructions, so the resulting bounds do not reflect the potential benefit of increasing depth. 

Despite these important developments, a key question is how to construct deep ReLU network estimators that achieve optimal prediction rates under norm-product constraints. To this end, we need to understand the expressive power of deep neural networks. Approximation theory quantifies this power by relating approximation error to network size \citep{anthony2009neural,yarotsky2017error,chen2019efficient,farrell2021deep,lu2021deep,siegel2023optimal,liu2025deep}. Using standard ReLU feedforward neural networks (FNNs) without a norm-product constraint, a large body of work has established approximation guarantees for smooth function classes  \citep{yarotsky2017error,lu2021deep,jiao2023deep,siegel2023optimal,liu2025deep}. A representative nearly optimal approximation rate is $W^{-2s/d}D^{-2s/d}$, up to logarithmic factors \citep{lu2021deep,jiao2023deep}. Here $s$ is the smoothness index of the target function, $d$ is the input dimension, and $W$ and $D$ denote the network width and depth, respectively. Moreover, \citet{gao2024convergence} established approximation bounds for Lipschitz functions using neural networks with direct Lipschitz control. However, using computable norm-product bounds for Lipschitz control imposes stronger restrictions on the network class and poses greater challenges for approximation. Therefore, whether the optimal approximation rates can be established within this constrained class of deep neural networks becomes one of the most important problems to be studied in this work.

To study this issue, we first recall a common approximation approach based on local Taylor expansions \citep{yarotsky2017error,lu2021deep,jiao2023deep}. The key objective here is to approximate a given smooth function using a carefully designed ReLU network. The corresponding network construction involves three main steps. The first step is localization. The input domain is partitioned into small grid cells, so that the target function can be well approximated by a Taylor polynomial locally around an associated grid point. Away from cell boundaries, a ReLU network approximates a step function that maps inputs to their corresponding grid points. The second step is local polynomial approximation. Each Taylor polynomial is a weighted sum of products. Weighted sums can be represented exactly by ReLU networks, so the main task is to approximate multiplication. Since multiplication can be expressed through squaring and weighted sums, $x^2$ serves as the most important and basic component for this approximation. The third step is coefficient encoding. Taylor coefficients at grid points are encoded in the network parameters and recovered approximately by point-fitting subnetworks.

Existing literature has shown that the above three-step construction can achieve the aforementioned nearly optimal rate if no constraint is imposed on the norm product. However, achieving this rate under a fixed norm-product constraint is very questionable. The key issue is whether the above three steps remain valid. In fact, imposing such a constraint would introduce serious challenges in each step. The first challenge is localization. In the constructions of \cite{lu2021deep} and \cite{jiao2023deep}, the grid resolution $K$ should be chosen to be of order  $W^{2/d}D^{2/d}$, up to logarithmic factors. As $K$ increases, the narrower transitions required for the target approximation rate make the grid-selection networks extremely steep near the cell boundaries. This leads to a diverging norm product. The second challenge is local polynomial approximation. A common approach uses repeated compositions of a piecewise linear sawtooth function to approximate the basic nonlinear component $x^2$ efficiently. Nevertheless, it can be verified that the resulting norm product also diverges to infinity as the approximation error decreases if this repeated sawtooth technique is used \citep{yarotsky2017error,jiao2023approximation}. The third challenge is coefficient encoding. Existing constructions use bit-extraction techniques to recover local Taylor coefficient information \citep{lu2021deep,siegel2023optimal}. The merit of this technique is that the coefficient information can be stored using a relatively small number of network parameters. However, the price paid is that the norm products of these bit-extraction networks also diverge as the encoding precision increases. Therefore, how to achieve accurate approximation under a fixed norm-product constraint becomes a problem of great importance for establishing optimal prediction rates.


In this paper, we develop a novel method for approximation and estimation with deep ReLU networks under norm-product control. Specifically, we start with the standard ReLU FNN class and define a constrained subclass using a novel split spectral norm product (SSNP), which is to be introduced in Section~\ref{subsec:NNclass}. This SSNP provides a computable upper bound on the Lipschitz constant of each network. Our main technical contribution is to develop new network construction techniques for approximation under a fixed SSNP constraint. Using these techniques, we show theoretically that sufficiently smooth H\"{o}lder functions can be approximated with uniform error of order $D^{-s/d}$, up to logarithmic factors. The SSNP bound remains fixed as the depth $D$ increases. Thus, these H\"{o}lder classes can be approximated arbitrarily well under fixed width and SSNP constraints. Next, we consider how to estimate the regression function using deep ReLU networks under SSNP control. We propose a penalized least squares estimator over the standard ReLU FNN class. The key idea is to penalize the SSNP in the empirical loss. With suitable choices of network size and penalty level, we establish a non-asymptotic bound on the expected squared prediction error. This bound matches the optimal rate $n^{-2s/(2s+d)}$, up to a logarithmic factor. The expected SSNP also remains bounded as the sample size and the network depth increase. Consequently, the expected Lipschitz constant of the resulting estimator also remains bounded. We further consider applications to two-stage learning and adversarial robustness. We show that the SSNP control bounds the effects of first-stage estimation errors and adversarial input perturbations on prediction. Extensive numerical studies show that the penalized estimator has lower input sensitivity than its unpenalized counterpart, with comparable prediction accuracy.

The remainder of the paper is organized as follows. Section~\ref{sec:pre} introduces the regression model and function classes. Section~\ref{sec:key} presents the three construction techniques for approximation under norm-product control. Section~\ref{sec:app} establishes the main approximation theorem and presents the supporting propositions. Section~\ref{sec:estimation} introduces the penalized least squares estimator, derives bounds on its expected prediction error and expected norm product, and studies applications to two-stage learning and adversarial robustness. Section~\ref{sec:simulation} presents numerical studies, and Section~\ref{sec:con} concludes the paper. 

\section{Preliminaries}\label{sec:pre}
In this section, we first introduce the basic setup of the nonparametric regression model. Then, we specify the function class of the target regression function and describe the neural network classes to be used in this paper. 

\subsection{Nonparametric regression model}\label{subsec:nonreg}
Let $\mathcal D_n=\{(X_i,Y_i):1\leq i\leq n\}$ be a sample of independent and identically distributed observations. Here $X_i=(X_{i1},\ldots,X_{id})^\top\in\mR^d$ is a $d$-dimensional feature vector, $Y_i\in\mR$ is the response, and $n$ is the total sample size. Throughout the paper, the covariate is assumed to be supported on $\mathcal X=[0,1]^d$. We consider the nonparametric regression model
\begin{equation}\label{equ:model}
	Y_i=f_0(X_i)+\varepsilon_i,
\end{equation}
where $f_0:[0,1]^d\to\mR$ is the unknown regression function and $\varepsilon_i$ is the random error satisfying $\E(\varepsilon_i\mid X_i)=0$ and $\var(\varepsilon_i\mid X_i)=\sigma_{\varepsilon}^2$. Next, we impose a smoothness condition on the true regression function $f_0$. Specifically, we consider an $s$-H\"{o}lder class. Write $\mathbb N_0=\{0,1,\ldots\}$ and $\mathbb N^+=\{1,2,\ldots\}$. For any multi-index $w=(w_1,\ldots,w_d)^{\top}\in\mathbb N_0^d$, write $\|w\|_1=\sum_{k=1}^d w_k$ and denote $\partial^w f= \partial^{w_1}\cdots\partial^{w_d}f(x)/ \partial x_1^{w_1}\cdots\partial x_d^{w_d}$ as the partial derivative of $f$ of order $w$. Here $\mathbb N_0^d$ is the set of $d$-dimensional vectors with non-negative integer components. For $s>0$, let $u=\lceil s\rceil-1$ and $r=s-u\in(0,1]$. Write $\|g\|_\infty=\sup_{x\in[0,1]^d}|g(x)|$. For any constant $B\geq 0$, define an $s$-H\"{o}lder class by
\begin{gather*}
	\mH^{s}\big([0,1]^d,B\big) \\
	= \bigg\{ f:[0,1]^{d}\to \mR :
	\max_{\lVert w \rVert_1 \le u} \|\partial^w f\|_{\infty} \le B,
	\max_{\lVert w \rVert_1 = u}\sup_{x\neq y}\frac{|\partial^w f(x)-\partial^w f(y)|}{\|x-y\|_2^r} \le B \bigg\},
\end{gather*}
We assume that the true regression function $f_{0}\in\mH^{s}\big([0,1]^d,B_0\big)$ for a fixed constant $B_0>0$.

\subsection{Neural network classes}\label{subsec:NNclass}

We consider feedforward neural networks with the rectified linear unit (ReLU) activation function $\sigma(t)=\max\{t,0\}$ \citep{nair2010rectified}. Next, for an integer $\mD\geq1$, define the $\ell$th layer transformation as $\phi_\ell(x)=W^{(\ell)}x+b^{(\ell)}$ with $W^{(\ell)}\in\mR^{p_\ell\times p_{\ell-1}}$ and $b^{(\ell)}\in\mR^{p_\ell}$ for $1\leq\ell\leq\mD+1$. Define $\sigma\odot x=(\sigma(x_1),\ldots,\sigma(x_q))^\top$ as the componentwise ReLU transformation for any vector $x=(x_1,\ldots,x_q)^\top\in\mR^q$. Then a network with $\mD$ hidden layers can be written as
\begin{equation}\label{equ:fnn}
	f_\Theta(x)
	=\phi_{\mD+1}\circ\sigma\odot\phi_{\mD}\circ\cdots\circ\sigma\odot\phi_1(x),
\end{equation}
where $\Theta=\{(W^{(\ell)},b^{(\ell)}):1\leq\ell\leq\mD+1\}$ denotes its parameter collection, with $p_0=d$ and $p_{\mD+1}=1$. Here $\circ$ represents the usual function composition. The total number of parameters is $\mS=\sum_{\ell=1}^{\mD+1}(p_{\ell-1}+1)p_\ell$. Moreover, define the network width as $\mW=\max\{p_0,\ldots,p_{\mD+1}\}$. Collect all these deep neural networks with $\mD$ hidden layers and width at most $\mW$ to obtain the feedforward neural network (FNN) function class $\mF_{\text{FNN}}(\mD,\mW)$. This is the standard ReLU feedforward architecture \citep{bauer2019deep,schmidt2020nonparametric,jiao2023deep,bhattacharya2024deep}. 

Next, we introduce a novel split spectral norm product (SSNP). Before giving its formal definition, we need to understand the potential issue of existing norm-product bounds. To fix ideas, consider a simple example with $Z_1(x)=1/2-\sigma(x-1/2)-\sigma(1/2-x)$ for $x\in[0,1]$. It can be represented by $Z_1(x)=W^{(2)}\sigma(W^{(1)}x+b^{(1)})+b^{(2)}$ with $W^{(1)}=(1,-1)^{\top}$, $b^{(1)}=(-1/2,1/2)^{\top}$, $W^{(2)}=(-1,-1)$ and $b^{(2)}=1/2$. It can be verified that $Z_1$ has Lipschitz constant $1$. However, the classical spectral norm product gives an upper bound of $2$ on its sensitivity to input perturbations. Moreover, the norm product defined in \cite{jiao2023approximation} equals $3$. Directly composing $m$ copies of the above representation of $Z_1$ gives a network with spectral norm product $2^m$. The norm product in \cite{jiao2023approximation} behaves similarly. In contrast, the represented function is still $Z_1$ and thus its Lipschitz constant remains $1$. This gap suggests that these two norm products can give overly conservative bounds on the Lipschitz constant. In this example, this conservativeness arises because the two units in each hidden layer cannot be active simultaneously. Those two norm products do not account for this restriction and hence give bounds that are not tight in this example. To address this problem, we develop the following SSNP, which incorporates this restriction and gives the exact bound $1$ for both the original and the composed representations.

We now define a subclass of $\mF_{\text{FNN}}(\mD,\mW)$ by constraining the SSNP of a specified network representation. Specifically, for a matrix $W\in\mR^{m\times p}$ and a vector $b\in\mR^m$, define
\begin{equation*}
	\mI(W,b) =\Big\{I\subseteq\{1,\ldots,m\}: (W_i,b_i)\neq-(W_j,b_j) \text{ for all }i\neq j\in I \Big\}.
\end{equation*}
Here $W_i$ is the $i$th row of $W$. Since two coordinates with opposite affine forms cannot be positive simultaneously, $\mI(W,b)$ contains every possible set of active coordinates of the ReLU map $u\mapsto\sigma(Wu+b)$ \citep{perugachi2021invertible}. For a network with representation $\Theta$, let $\mI_\ell=\mI(W^{(\ell)},b^{(\ell)})$ for $1\leq\ell\leq\mD$. Set $\mI_0=\{\{1,\ldots,d\}\}$ and $\mI_{\mD+1}=\{\{1\}\}$. For each layer, define $\|W^{(\ell)}\|_{\text{ssnp},\Theta} = \max_{I\in\mI_\ell,J\in\mI_{\ell-1}} \|W^{(\ell)}_{I,J}\|_{\text{op}} $ for every $1\leq\ell\leq\mD+1$, where $W^{(\ell)}_{I,J}$ is the submatrix formed by the rows indexed by $I$ and the columns indexed by $J$, and $\|\cdot\|_{\text{op}}$ denotes the spectral norm. The norm of a submatrix with no rows or no columns is understood to be zero. The row and column selections account for the active coordinates in the current and preceding layers, respectively. Thus, a restriction on the outputs of one layer is also taken into account in the input columns of the next layer. Define 
\begin{equation}\label{equ:rhonorm}
	\rho(f_\Theta) =\prod_{\ell=1}^{\mD+1}	\|W^{(\ell)}\|_{\text{ssnp},\Theta}.
\end{equation}
Since the output is scalar, we have $\|W^{(\mD+1)}\|_{\text{ssnp},\Theta} =\max_{J\in\mI_\mD} \|W^{(\mD+1)}_{\{1\},J}\|_{\text{op}}$. For the representation of $Z_1$ above, we have $\|W^{(1)}\|_{\text{ssnp},\Theta}=\|W^{(2)}\|_{\text{ssnp},\Theta}=1$. This leads to $\rho(Z_1)=1$.
Let $\text{Lip}(f)=\sup_{x\neq y\in[0,1]^d}|f(x)-f(y)|/\|x-y\|_2$ denote the Lipschitz constant of a function $f$ with respect to the Euclidean norm. The SSNP defined in \eqref{equ:rhonorm} satisfies $\text{Lip}(f_\Theta)\leq\rho(f_\Theta)$, as proved in Lemma 14 of Appendix C. Throughout, $\rho(f_\Theta)$ denotes the SSNP computed for the specified network representation $\Theta$. Different representations of the same function may have different norm products. For any $\mG>0$, define the SSNP constrained neural network (SSNN) class by  $\mF_{\text{SSNN}}(\mD,\mW,\mG)=\big\{ f_\Theta\in\mF_{\text{FNN}}(\mD,\mW): \rho(f_\Theta)\leq\mG \big\}$.

\section{Three key techniques}\label{sec:key}

To address the challenges of localization, local polynomial approximation, and coefficient encoding discussed in Section~\ref{Intro}, we develop three novel techniques for approximation in the SSNN class. Our construction builds on local Taylor approximation, with the SSNP controlled throughout the network construction. These techniques provide the important ingredients for the main approximation theorem in Section~\ref{sec:app}.

%

\subsection{SSNP controlled approximation of $x^2$}
\label{subsec:square-approx}

We begin with the approximation of $x^2$ with controlled SSNP. As described in Section~\ref{Intro}, $x^2$ is the most fundamental nonlinear component to be approximated. The standard FNN approximation of $x^2$ is based on sawtooth refinements \citep{yarotsky2017error,lu2021deep}. Specifically, define $T_1:[0,1]\to[0,1]$ by $T_1(x)=2x$ for $x\in[0,1/2]$ and $T_1(x)=2(1-x)$ for $x\in(1/2,1]$, and set $T_i(x)=T_{i-1}\circ T_1(x)$ for $i\geq 2$. Then, each composition refines the resolution in the sense that $T_i$ has $2^{i-1}$ sawteeth. Let $f_m(x)$ be the piecewise linear interpolant of $x^2$ on the grid $\{j2^{-m}:j=0,1,\ldots,2^m\}$. For $x\in[0,1]$, we can obtain that
\begin{equation}\label{equ:fs}
	f_m(x)=x-\sum_{i=1}^m \frac{T_i(x)}{2^{2i}}.
\end{equation}
In this regard, the interpolation error satisfies $ \sup_{x\in[0,1]}\big|f_m(x)-x^2\big|\leq 2^{-2(m+1)}$. Therefore, the approximation error decays exponentially fast with respect to the refinement level $m$. However, this efficient approximation comes at a heavy cost in terms of spectral-norm budget. Specifically, the basic sawtooth function admits the standard ReLU representation  $T_1(x)=2\sigma(x)-4\sigma(x-1/2)+2\sigma(x-1)$ for $x\in[0,1]$. A direct computation shows that a single $T_1$ block has spectral-norm product $6\sqrt{2}$. Composing $m$ copies of this unscaled block gives a layerwise norm-product bound $(6\sqrt{2})^m$ for $T_m$. This bound grows exponentially fast with $m$. Instead, it is totally due to this particular choice of sawtooth approximation technique and has little to do with the target function $x^2$. Then, how to develop a novel technique to replace sawtooth method with better Lipschitz properties becomes the most important problem.

To this end, we develop a new technique. Instead of working with the original sawtooth function $T_i(x)$, we consider its normalized version as $Z_i(x)=T_i(x)/2^i$. The functions $Z_i$ and $T_i$ have the same breakpoints, but each $Z_i$ is $1$-Lipschitz. Then, the expansion in \eqref{equ:fs} becomes $f_m(x)=x-\sum_{i=1}^m Z_i(x)/2^i$. More importantly, the normalized functions satisfy 
\begin{equation*}
	Z_{i+j}=S_j^{-1}\circ Z_i\circ S_j\circ Z_j
\end{equation*}
for any $i,j\geq 1$, where $S_j(x)=2^jx$ and $ S_j^{-1}(x)=2^{-j}x$. Here $S_j(x)$ allows an intermediate amplification by a factor $2^j$, while $S_j^{-1}(x)$ rescales it back by $2^{-j}$. Since $\|S_j\|_{\text{op}}\|S_j^{-1}\|_{\text{op}}=1$, this amplification does not increase the overall norm product. The basic normalized function has the ReLU representation $Z_1(x)=1/2-\sigma(x-1/2)-\sigma(1/2-x)$ on $[0,1]$. This scalar function is $1$-Lipschitz. The remaining cost comes only from the initialization and the summation state. This leads to the following SSNP controlled approximation result for $x^2$ in Lemma~\ref{lem:x2}, whose detailed proof is given in Appendix C. By Lemma~\ref{lem:x2}, we know that the classical exponential approximation rate for $x^2$ can be retained with a fixed SSNP, which is upper bounded by $ \sqrt{2e}$. 
\begin{lemma}\label{lem:x2}
	For any $D \in \mathbb{N}^+$, there exists a function $\phi\in\mF_{\text{\normalfont SSNN}}(D,3, \sqrt{2e})$ such that $\sup_{x\in[0,1]}\big|\phi(x) - x^2\big| \leq 2^{-D}$.
\end{lemma}

\subsection{Continuous patching of local approximants}\label{subsec:local-patching}

Next, we consider how to patch local polynomial approximants into a global network with a controlled norm product. A Taylor construction produces polynomial pieces on local regions. For each input point, one must specify how these pieces are then combined. As discussed in Section~\ref{Intro},  standard FNN constructions often use hard selectors to choose the active local piece \citep{schmidt2020nonparametric,lu2021deep,jiao2023deep}. Specifically, for an integer $K\ge 1$, define $\mI_K= \{0,\ldots,K-1\}^d$ and $\overline{\mI}_K=\{0,\ldots,K\}^d$. Partition $[0,1]^d$ into cubes $Q_{k,K} =\prod_{j=1}^d\big[k_j/K, (k_j+1)/K\big)$, indexed by $k=(k_j)\in \mI_K$ with the usual boundary convention on the last interval. This leads to $\bigcup_{k\in \mI_K}Q_{k,K} =[0,1]^d$. Then, for each $k\in\overline{\mI}_K$, let
\begin{equation*}
	T_{k,K}f(x)=\sum_{\|\alpha\|_1\leq u}\frac{\partial^\alpha f(x_{k,K})}{\alpha!}(x-x_{k,K})^\alpha
\end{equation*}
be the degree-$u$ Taylor polynomial of $f$ at $x_{k,K}=k/K$.  Here $\alpha=(\alpha_j)$ is a multi-index, $\alpha!=\alpha_1!\cdots\alpha_d!$, and $(x-x_{k,K})^\alpha=\prod_{j=1}^d\big(x_j-k_j/K\big)^{\alpha_j}$. A hard-selector construction defines the global piecewise-polynomial approximation by 
\begin{equation*}
	T_K^{\rm hard}f(x)= \sum_{k\in\mI_K}I\big(x\in Q_{k,K}\big)T_{k,K}f(x),
\end{equation*}
where $I(\cdot)$ is an indicator function. For each input $x$, one cell selector is active, and the corresponding Taylor polynomial is used. However, this construction need not be continuous across grid boundaries and thus does not directly yield an SSNN approximation.

A natural continuous alternative is to replace hard selectors with some soft ones. For example, a standard choice is the partition-of-unity (PoU) construction \citep{yarotsky2017error,jiao2023approximation,yang2024nonparametric}. Let $\psi(t)=\sigma(1-|t|)$. For each $k\in\overline{\mI}_K$, define $\Psi_{k,K}(x)=\prod_{j=1}^d\psi(Kx_j-k_j)$. We then have $\sum_{k\in\overline{\mI}_K}\Psi_{k,K}(x)=1$ for any $x\in[0,1]^d$. Then, a partition-of-unity approximation is given by 
\begin{equation*}
	T^{\rm PoU}_Kf(x)=\sum_{k\in\overline{\mI}_K} \Psi_{k,K}(x)T_{k,K}f(x).
\end{equation*}
Define $A_j(x)=\big\{m\in\{0,\ldots,K\}: \psi(Kx_j - m)>0\big\}$. Since $\psi(t)=0$ for $|t|\ge 1$, we know that $|A_j(x)|\leq 2$ for every $1\leq j\leq d$. Moreover, we have $\Psi_{k,K}(x)>0$ if and only if $k_j\in A_j(x)$ for all $j=1,\ldots, d$. This implies that  $T_K^{\rm PoU}f$ is a continuous local average of at most $2^d$ neighboring Taylor polynomials. Therefore, $T_K^{\rm PoU}f$ preserves the approximation accuracy of the local Taylor expansion. The remaining issue is how to approximate it effectively by neural networks. In standard PoU-based FNN constructions, the partition weights are constructed first and then combined with the Taylor monomials through product networks. Each factor $t\mapsto\psi(Kt-k_j)$ has Lipschitz constant $K$. Thus, constructing these factors separately introduces norm-product contributions that grow with $K$ and must be compensated elsewhere in the construction.

To address this issue, we develop a new continuous patching technique. The key observation is that directly approximating the partition weight $\Psi_{k,K}$ by a neural network is unnecessarily expensive in the norm product. Instead, we might consider approximating its localized Taylor product $\Psi_{k,K}(x)(x-x_{k,K})^\alpha$ as a whole. This distinction is important because the Lipschitz constant of $\Psi_{k,K}$ is of order $K$. Moreover, for a fixed multi-index $\alpha$, the corresponding PoU component takes the form $\sum_{k\in\overline{\mI}_K}c_{k,\alpha}\Psi_{k,K}(x)(x-x_{k,K})^\alpha$ with $c_{k,\alpha}=\partial^{\alpha}f(x_{k,K})/\alpha!$ and $|\overline{\mI}_K|=(K+1)^d\leq 2^dK^d$. Thus, under a parallel aggregation over $k\in\overline{\mI}_K$, the number of localized terms contributes a factor of order $K^d$ to the norm product accounting. On the support of $\Psi_{k,K}$, we have $|x_j-k_j/K|\leq K^{-1}$ for every $1\leq j\leq d$, and hence $\big|(x-x_{k,K})^\alpha\big| \leq K^{-r}$ for $r=\|\alpha\|_1$. By approximating the localization weight and the Taylor factor jointly, the small factor $K^{-r}$ is incorporated before aggregation over the grid. Consequently, the resulting norm-product upper bound is of order $K^{d+1-r}$, up to constants independent of $K$. For $r\ge d+1$, this bound is uniform in $K$, so the joint approximation of localized Taylor products already suffices to control the product spectral-norm budget. 

Nevertheless, if we unfortunately have $r=0$, then no Taylor-scale factor is available. In this case, we have $(x-x_{k,K})^\alpha=1$, so the localized Taylor product reduces to $\Psi_{k,K}(x)$. We treat this case by difference refinement in the next subsection. For $1\leq r\leq d$, we combine the localized-product construction with differences of the derivative coefficients.

\subsection{Difference refinement by localized interpolation errors}\label{subsec:difference}

We now focus on the most challenging case with $r=0$. In this case, the approximation target becomes $T_{K,0}^{\rm PoU}f(x)= \sum_{k\in\overline{\mI}_K} f(x_{k,K})\Psi_{k,K}(x)$. As discussed before, directly approximating this sum by a neural network can lead to a norm-product bound that grows with $K$. However, we find that this problem can be resolved by a difference refinement technique using local interpolation errors of $f$. Specifically, assume $K=2^J$ for some fixed integer $J>0$. For $0\le j\le J$, let $K_j=2^j$, $h_j=K_j^{-1}$, and $x_{\ell,K_j}=\ell/K_j$ for $\ell\in\overline{\mI}_{K_j}=\{0,\ldots,K_j\}^d$. Let $I_jf(x)=T_{K_j,0}^{\rm PoU}f(x)$ denote the PoU interpolant of $f$ at resolution $K_j$. Since $K=K_J$, we have $T_{K,0}^{\rm PoU}f=I_Jf$. Define a difference $\Delta_jf=I_{j+1}f-I_jf$ for $j=0,\ldots,J-1$. This leads to 
\begin{equation*}
	T_{K,0}^{\rm PoU}f=I_Jf=I_0f+\sum_{j=0}^{J-1}\Delta_jf.
\end{equation*}
Instead of approximating $T_{K,0}^{\rm PoU}f$ in one step, we can start from $I_0f$ and add the corrections $\Delta_jf$ successively. Note that each $\Delta_jf$ is the difference between the interpolants at two neighboring resolutions. They are controlled by the smoothness of $f$ and thus provide the sufficient small factors needed in the norm product calculation.

For illustration purposes, we present here the construction details for $d=1$, and give the higher dimensional cases in Appendix A.3. For $d=1$, the target is the linear interpolant of $f$ on $\{k/K:0\le k\le K\}$. More precisely, for $x\in Q_{k,K}=[k/K,(k+1)/K]$, we have
\begin{equation*}
	T_{K,0}^{\rm PoU}f(x)=f(x_{k,K})\Psi_{k,K}(x)+f(x_{k+1,K})\Psi_{k+1,K}(x).
\end{equation*}
Thus, $T_{K,0}^{\rm PoU}f=I_Jf$ is the piecewise linear interpolant at resolution level $K=K_J$. Moreover, $I_jf$ is the piecewise linear interpolant on $\{x_{j,k}:0\le k\le K_j\}$ with $x_{j,k}=kh_j$. At level $j$, we have $\Delta_jf$ given by
\begin{equation*}
	\Delta_jf(x)=\sum_{k=0}^{K_j-1}\delta_{j,k}\Psi_{2k+1,K_{j+1}}(x),
\end{equation*}
where $\delta_{j,k}=f(x_{j+1,2k+1})-\big\{f(x_{j,k})+f(x_{j,k+1})\big\}/2$ is the interpolation error of $f$. For $f\in\mH^s([0,1],B_0)$ with $s>3/2$, direct computation leads to $|\delta_{j,k}|\leq CB_0h_j^{\min\{s,2\}}$ for some constant $C>0$. Then, the remaining task is to show that this difference refinement can be constructed by a ReLU SSNN. The detailed construction and norm-product calculation are given in the case $d=1$ of the proof of Proposition 4.2 in Appendix A.3. This one-dimensional example illustrates how the decay of the interpolation errors can be used to keep the SSNP independent of $K$. This is the role of difference refinement.

%
%

\section{Approximation theory}\label{sec:app}

This section establishes the approximation theory for ReLU SSNNs. The key issue is how a fixed bound on the SSNP affects the approximation power of ReLU FNNs. We first state the main approximation theorem and present its proof sketch. We then compare the result with existing approximation theory and collect the supporting propositions used in the construction.

\subsection{Main approximation result}\label{subsec:mainapp}

We start with the main approximation theorem for ReLU SSNNs. Theorem~\ref{thm:lnn-holder-approx} gives a uniform approximation bound for ReLU SSNNs over $\mH^s([0,1]^d,B_0)$ with $s>d/2+1$. The approximation error is of order $D^{-s/d}$ for networks of depth $O\{D\log_2(2D)\}$. The width and the SSNP bound do not increase with $D$. A proof sketch follows the theorem. The detailed proof is given in Appendix A.1.
\begin{theorem}\label{thm:lnn-holder-approx}
	Let $f\in\mH^s([0,1]^d,B_0)$ with $s>d/2+1$ and $B_0\geq0$. There exist constants $C_{d,s}\geq1$, $W_{d,s}^{\mathrm{app}}\in\mathbb N^+$, and $G_{d,s}^{\mathrm{app}}\geq1$ such that the following holds. For any $D\in\mathbb N^+$, there exists a ReLU network $\Phi_D\in\mF_{\text{SSNN}}\big(L_D,W_{d,s}^{\mathrm{app}},\max\{1,B_0\}G_{d,s}^{\mathrm{app}}\big)$ with $L_D=\lceil C_{d,s}D\log_2(2D)\rceil$ satisfying
	\begin{equation}\label{equ:app-main-error}
		\big\|f-\Phi_D\big\|_{\infty} \leq C_{d,s}B_0D^{-s/d}.
	\end{equation}
\end{theorem}

\begin{proof}[Proof sketch]
The proof follows the key construction in the previous section. We first consider $B_0=1$ and set $u=\lceil s\rceil-1$. Choose $J=\max\big\{1,\lfloor \log_2D/d\rfloor\big\}$, $ K=2^J$ and $ h_K=1/K$. We then have $D^{1/d}/2\leq K\leq2D^{1/d}$. We first approximate $f$ by a grid-based Taylor quasi-interpolant $\mT_Kf\coloneq T_K^{\rm PoU}f$ and then approximate $\mT_Kf$ by a ReLU SSNN $\Phi$. The construction can be summarized as
\begin{equation*}
	f\xrightarrow[\text{partition of unity}]{\text{local Taylor expansion}}\mT_Kf
	\xrightarrow[\text{sequential approximation}]{\text{ReLU SSNN}}\Phi.
\end{equation*}
Writing $\|g\|_\infty=\sup_{x\in[0,1]^d}|g(x)|$, the triangle inequality gives
\begin{equation*}
	\|f-\Phi\|_\infty\leq\|f-\mT_Kf\|_\infty+\|\mT_Kf-\Phi\|_\infty.
\end{equation*}
Therefore, it remains to bound the Taylor approximation error and the SSNN approximation error. This is accomplished in the following two steps.

\sc{Step 1. Grid partition and Taylor expansion.}\normalfont$\ $
Recall that $\mI_K=\{0,1,\ldots,K-1\}^d$ and $\overline{\mI}_K=\{0,1,\ldots,K\}^d$. For each $\tau\in\overline{\mI}_K$, let $x_{\tau,K}=\tau/K$. For $k=(k_j)\in\mI_K$, define $Q_{k,K}=\prod_{j=1}^d[k_j/K,(k_j+1)/K)$, including the right endpoint in each coordinate interval ending at one. Consider the continuous partition of unity on this grid. Recall that $\psi(t)=\sigma(1-|t|)$ and $\Psi_{\tau,K}(x)=\prod_{j=1}^d\psi(Kx_j-\tau_j)$ for $\tau\in\overline{\mI}_K$. We have $0\leq\Psi_{\tau,K}(x)\leq1$, $\sum_{\tau\in\overline{\mI}_K}\Psi_{\tau,K}(x)=1$ for every $x\in[0,1]^d$, and $\Psi_{\tau,K}(x)=0$ unless $\|x-x_{\tau,K}\|_\infty\leq h_K$. Thus, the functions $\{\Psi_{\tau,K}:\tau\in\overline{\mI}_K\}$ form a localized continuous partition of unity. In particular, if $x\in Q_{k,K}$, then the only possibly nonzero weights are $\Psi_{k+\nu,K}(x)$ with $\nu\in\{0,1\}^d$.

For each $\tau\in\overline{\mI}_K$, define the local Taylor polynomial
\begin{equation*}
	T_{\tau,K}f(x)=\sum_{\|\alpha\|_1\leq u}\frac{\partial^\alpha f(x_{\tau,K})}{\alpha!}(x-x_{\tau,K})^\alpha,
\end{equation*}
where $\alpha\in\mathbb N_0^d$, $\alpha!=\prod_{j=1}^d\alpha_j!$, and $z^\alpha=\prod_{j=1}^dz_j^{\alpha_j}$. The grid-based Taylor quasi-interpolant is defined by $\mT_Kf(x)=\sum_{\tau\in\overline{\mI}_K}\Psi_{\tau,K}(x)T_{\tau,K}f(x)$. Equivalently, for $x\in Q_{k,K}$,
\begin{equation*}
	\mT_Kf(x)=\sum_{\nu\in\{0,1\}^d}\Psi_{k+\nu,K}(x)T_{k+\nu,K}f(x).
\end{equation*}
The partition of unity gives $f(x)-\mT_Kf(x)=\sum_{\nu\in\{0,1\}^d}\Psi_{k+\nu,K}(x)\{f(x)-T_{k+\nu,K}f(x)\}$. Since $\|x-x_{k+\nu,K}\|_\infty\leq h_K$, the Taylor remainder bound gives $|f(x)-T_{k+\nu,K}f(x)|\leq C_{d,s}h_K^s$; see Lemma A.8 of \citet{petersen2018optimal}. Using $\sum_{\nu\in\{0,1\}^d}\Psi_{k+\nu,K}(x)=1$, we obtain
\begin{equation*}
	\big\|f-\mT_Kf\big\|_\infty\leq C_{d,s}K^{-s}.
\end{equation*}

\sc{Step 2. SSNN approximation.}\normalfont$\ $
We now approximate $\mT_Kf$ by a ReLU SSNN. For $x\in Q_{k,K}$, recall that
\begin{equation*}
	\mT_Kf(x)=\sum_{\nu\in\{0,1\}^d}\sum_{\|\alpha\|_1\leq u}\frac{\partial^\alpha f(x_{k+\nu,K})}{\alpha!}\Psi_{k+\nu,K}(x)(x-x_{k+\nu,K})^\alpha.
\end{equation*}
The factor $\Psi_{k+\nu,K}(x)$ localizes the Taylor monomial $(x-x_{k+\nu,K})^\alpha$ around the vertex $x_{k+\nu,K}$. Thus, we approximate $\Psi_{k+\nu,K}(x)(x-x_{k+\nu,K})^\alpha$ as a whole block. To understand its scaling, let $u_K(x)=(u_{K,j}(x))\in[0,1]^d$ be the normalized local coordinate of $x$ in $Q_{k,K}$, where $u_{K,j}(x)=Kx_j-k_j$. Let $\ell_0(t)=1-t$ and $\ell_1(t)=t$. Then
\begin{equation*}
	\Psi_{k+\nu,K}(x)(x-x_{k+\nu,K})^\alpha
	=h_K^{\|\alpha\|_1}\prod_{j=1}^d\ell_{\nu_j}(u_{K,j}(x))\{u_{K,j}(x)-\nu_j\}^{\alpha_j}.
\end{equation*}
Therefore, a Taylor term of order $r=\|\alpha\|_1$ carries an intrinsic factor $h_K^r$. This factor is absent for $r=0$, is not sufficient by itself for the direct localization construction when $1\leq r\leq d$, and compensates for the localization cost when $r\geq d+1$. This leads to three separate treatments.

Specifically, for $r=0,1,\ldots,u$ and $x\in Q_{k,K}$, define
\begin{equation*}
	\mT_{K,r}f(x)=\sum_{\|\alpha\|_1=r}\sum_{\nu\in\{0,1\}^d}\frac{\partial^\alpha f(x_{k+\nu,K})}{\alpha!}\Psi_{k+\nu,K}(x)(x-x_{k+\nu,K})^\alpha.
\end{equation*}
We then have $\mT_Kf=\sum_{r=0}^u\mT_{K,r}f$. We further group these terms as
\begin{equation*}
	\mT_Kf=\mT_{K,0}f+\sum_{r=1}^{\min\{u,d\}}\mT_{K,r}f+\sum_{r=d+1}^u\mT_{K,r}f,
\end{equation*}
where an empty sum is understood as zero. To approximate $\mT_Kf$ by a ReLU SSNN, we carry out the following four tasks.
\begin{itemize}
	\item[Task 1.] Construct a basic SSNN approximation of the monomial $x^\alpha$ for each multi-index $\alpha$ with $\|\alpha\|_1\leq u$. Starting from Lemma~\ref{lem:x2}, the product construction gives Proposition~\ref{prop:xalpha}.
	\item[Task 2.] Construct an SSNN approximation of $\mT_{K,0}f$. This component corresponds to the continuous interpolation of neighboring function values and has no intrinsic small Taylor factor. It is handled by the multiscale difference refinement in Section~\ref{subsec:difference}, leading to Proposition~\ref{prop:zero}.
	\item[Task 3.] Construct an SSNN approximation of the low-order Taylor corrections $\mT_{K,r}f$ for $1\leq r\leq\min\{u,d\}$. These terms are handled by combining localized products with differences of derivative coefficients and componentwise rescaling, as stated in Proposition~\ref{prop:fixed-intermediate}.
	\item[Task 4.] Construct an SSNN approximation of the high-order Taylor corrections $\mT_{K,r}f$ for $d+1\leq r\leq u$. These terms are handled by the continuous patching construction in Section~\ref{subsec:local-patching}. Proposition~\ref{prop:fixed-high} treats each multi-index $\alpha$ separately, and summing over $\|\alpha\|_1=r$ gives the required approximation of $\mT_{K,r}f$.
\end{itemize}
Propositions~\ref{prop:zero}, \ref{prop:fixed-intermediate}, and~\ref{prop:fixed-high} in Section~\ref{subsec:suppprops} specify the component networks in Tasks~2--4. For the zero- and low-order terms, set depth $\lceil C_{d,s}K^d\rceil$. For each high-order term, set depth $\lceil C_{d,s}K^d\log_2(2K)\rceil$. With $C_{d,s}$ sufficiently large, each component has approximation error at most $C_{d,s}K^{-s}$. Its width and SSNP are bounded independently of $K$. Denote these networks by $\phi_1,\ldots,\phi_m$, where $m$ depends only on $d$ and $s$. Their sum $\Phi=\sum_{j=1}^m\phi_j$ satisfies
\begin{equation*}
	\|\mT_Kf-\Phi\|_\infty\leq C_{d,s}K^{-s}.
\end{equation*}

By the parallel construction in Appendix A.1, $\Phi$ has width at most $W_{d,s}^{\mathrm{app}}$ and satisfies $\rho(\Phi)\leq\sum_{j=1}^m\rho(\phi_j)\leq G_{d,s}^{\mathrm{app}}$. Its depth is at most $C_{d,s}K^d\log_2(2K)\leq C_{d,s}D\log_2(2D)$. Combining the preceding error bound with Step~1 gives
\begin{equation*}
	\|f-\Phi\|_\infty\leq C_{d,s}K^{-s}\leq C_{d,s}D^{-s/d},
\end{equation*}
which proves the result for $B_0=1$. For $B_0>0$, apply the same construction to $f/B_0$. Then, $B_0\Phi$ has the same depth and width as $\Phi$. It satisfies $\|f-B_0\Phi\|_\infty\leq C_{d,s}B_0D^{-s/d}$ and $ \rho(B_0\Phi)\leq\max\{1,B_0\}G_{d,s}^{\mathrm{app}}$. For $B_0=0$, we use the zero network.

\end{proof}

\subsection{Comparison with existing approximation bounds}\label{subsec:compare}

Theorem~\ref{thm:lnn-holder-approx} shows that approximation accuracy can improve with depth while both the width and the SSNP $\rho(f_\Theta)$ remain uniformly bounded. We compare this result with existing approximation bounds. Throughout this comparison, $D$ is the approximation parameter in Theorem~\ref{thm:lnn-holder-approx}, and $\mW$ and $\mD$ denote network width and depth, respectively.

For standard ReLU FNNs, \cite{yarotsky2017error} obtains uniform error of order $D^{-s/d}$ on the unit ball of $W^{s,\infty}([0,1]^d)$ for integer $s$. Here $W^{s,\infty}$ denotes the Sobolev space of functions with essentially bounded weak derivatives up to order $s$. The networks use $O\{D\log(2D)\}$ nonzero parameters and have depth $O\{\log(2D)\}$. For integer $s>d/2+1$, our construction obtains the same orders of approximation error and parameter count, with fixed width and a bounded norm product. In terms of width and depth, \cite{lu2021deep} obtain uniform bounds of order $\mW^{-2s/d}\mD^{-2s/d}$ up to logarithmic factors for smooth functions. \cite{jiao2023deep} give the corresponding bound for H\"{o}lder functions. \cite{gao2024convergence} also established approximation bounds for Lipschitz functions using deep ReLU networks with uniformly bounded Lipschitz constants. These existing approximation results are established without norm-product constraints.

For norm-constrained ReLU networks, \cite{jiao2023approximation} establish approximation bounds under a product constraint based on induced $\ell_\infty$ matrix norms, including hidden-layer biases. Their construction combines local Taylor polynomials through a continuous partition of unity with weights that sum to one \citep{yarotsky2017error}. They also use an integral representation to approximate $x^2$ by a weighted sum of ReLU functions. With grid resolution of order $D^{1/d}$, their construction achieves uniform error of order $D^{-s/d}$ with width of order $D^{(2d+s)/(2d)}$ and a norm budget of order $D^{(d+1)/d}$. Thus, both the width and the norm budget increase as the approximation error decreases. Their stated upper bound does not decrease with depth when the norm budget is fixed.

Joint control of approximation accuracy, width and parameter norms is provided by \citet{yang2024nonparametric,yang2025optimal} for shallow ReLU networks. These results are obtained under a variation-norm constraint. For H\"older functions on the unit Euclidean ball with $0<s<(d+3)/2$, they obtain uniform error of order $\mW^{-s/d}$ with a norm budget of order $\mW^{(d+3-2s)/(2d)}$. For $s>(d+3)/2$, the norm budget can remain fixed, and the error bound is of order $\mW^{-(d+3)/(2d)}$. Thus, sufficiently smooth functions can be approximated arbitrarily well under a fixed norm budget by increasing width. The extension to deep networks in \citet{yang2025optimal} gives an upper bound in terms of width and the norm budget. This bound does not decrease with depth.

The norm constraints in these works differ from the SSNP used here. Theorem~\ref{thm:lnn-holder-approx} complements these results by providing a depth-dependent approximation bound for every $s>d/2+1$, with both the width and the SSNP uniformly bounded.

\subsection{Supporting propositions}\label{subsec:suppprops}

We now state the four propositions used in the proof sketch of Theorem~\ref{thm:lnn-holder-approx}. Their proofs are given in Appendices A.2--A.5, respectively. First, we construct monomial approximations from Lemma~\ref{lem:x2} by repeated multiplication. These approximations are used to construct the localized Taylor terms. Proposition~\ref{prop:xalpha} shows that, for a fixed degree bound, the approximation error decreases exponentially with depth, while the width and norm budget remain fixed. Monomials of degree at most one are affine and can be represented exactly.
\begin{proposition}[Task 1. SSNN approximation of monomials]\label{prop:xalpha}
	Assume $P(x)=x^\alpha=x_1^{\alpha_1}x_2^{\alpha_2}$ $\cdots x_d^{\alpha_d}$ for some $\alpha\in\mathbb N_0^d$ with $\|\alpha\|_1\le k$, where $ k\geq 2$ is an integer. For any $D\in\mathbb N^+$, let $m \coloneq \lceil \log_2 k\rceil$ and $D_k:=D+\left\lceil 2^{-1}\log_2\!\big\{90(2.1^{m}-1)\big\}\right\rceil $. Then, there exists $\phi\in\mF_{\text{SSNN}}\big(mD_k,\allowbreak
	\max\{d,10k\},\allowbreak \sqrt{k}(6\sqrt{2e})^m\big)$ such that
	\begin{equation*}
		\sup_{x\in[0,1]^d} \big|\phi(x)-P(x)\big| \le 9\big(2.1^m-1\big)2^{-2D_k}.
	\end{equation*}
\end{proposition}

Second, we approximate the zeroth-order component $\mT_{K,0}f$ using the difference refinement described in Section~\ref{subsec:difference}.  Proposition~\ref{prop:zero} provides the SSNN approximation of the continuous interpolation of neighboring function values. Taking the depth to be a sufficiently large constant multiple of $K^d$ makes the error at most $C_{d,s}K^{-s}$, while the width and norm product remain independent of $K$.
\begin{proposition}[Task 2. Approximation of $\mT_{K,0}f$]\label{prop:zero}
	Fix $d\in\mathbb N^+$ and $s>d/2+1$. There exist constants $A_d\geq1$, $c_d>0$, $C_{d,s}>0$, and $G_{0,d,s}\geq1$ such that the following holds. Let $J\in\mathbb N^+$, $K=2^J$, and $f\in\mH^s([0,1]^d,1)$. For every integer $L\geq A_dK^d$, there exists a ReLU network $\Phi_{0,J,L}\in\mF_{\text{\normalfont SSNN}}(L,W_{0,d},G_{0,d,s})$, with $W_{0,d}=12d2^d$, satisfying
	\begin{equation*}
		\sup_{x\in[0,1]^d}\big|\Phi_{0,J,L}(x)-\mT_{K,0}f(x)\big|
		\leq C_{d,s}2^{-c_dL/(J+1)^d}.
	\end{equation*}
\end{proposition}

Third, we approximate the low-order Taylor corrections $\mT_{K,r}f$ for $1\leq r\leq\min\{u,d\}$. Compared with the zeroth-order component, they contain the intrinsic factor $h_K^r$ from the Taylor monomial. However, this factor is still not sufficient to control all localization costs by a direct product construction. We therefore use a support-adapted SSNN construction that combines localized products with differences of derivative coefficients and componentwise rescaling. Proposition~\ref{prop:fixed-intermediate} gives the resulting approximation of $\mT_{K,r}f$.  For $d\in\mathbb N^+$ and $s>d/2+1$, set $m_d=\lceil\log_2(2d)\rceil$, $N_d=\binom{2d-1}{d-1}$, $\kappa=\min\{1/8,(s-d/2-1)/(4d)\}$, and $Q_{d,s}=2+(1-2^{-\kappa})^{-1}$. Define
\begin{equation*}
 G_d=2^{d/2}4^d\sqrt{2d}\,(6\sqrt{2e})^{m_d}\exp\left\{(d+1)4^{d-1}+\frac{5d-2}{4}\right\}.
\end{equation*}
Let $W_d=20d4^d+8d+2$ and
\begin{equation*}
	G_{d,s}=
	\begin{cases}
		24e^{7/4},&d=1,\\
		\exp\{2^dN_dG_dQ_{d,s}^{d-1}\},&d\geq2.
	\end{cases}
\end{equation*}
\begin{proposition}[Task 3. Approximation of $\mT_{K,r}f$]\label{prop:fixed-intermediate}
	Fix $d\in\mathbb N^+$ and $s>d/2+1$. There exist constants $A_d\geq1$ and $c_d>0$ such that the following holds. Let $J\in\mathbb N^+$, $K=2^J$, and $f\in\mH^s([0,1]^d,1)$. For every integer $r$ satisfying $1\leq r\leq\min\{d,\lceil s\rceil-1\}$ and every integer $L\geq A_dK^d$, there exists a ReLU network $\Phi_{r,J,L}\in\mF_{\text{\normalfont SSNN}}(L,W_d,G_{d,s})$ satisfying
	\begin{equation*}
		\sup_{x\in[0,1]^d}\big|\Phi_{r,J,L}(x)-\mT_{K,r}f(x)\big|
		\leq2^{3d+2}N_dK^{-r}2^{-c_dL/(J+2)^{d-1}}.
	\end{equation*}
\end{proposition}

Finally, we approximate the high-order Taylor corrections, corresponding to $d+1\leq r\leq u$. In this regime, each localized Taylor basis carries the intrinsic factor $h_K^r$, which is sufficiently small to compensate for the localization cost. Therefore, the high-order terms can be handled by rescaled localized polynomial and product networks. For each $\alpha\in\mathbb N_0^d$, with $d+1\leq\|\alpha\|_1\leq u$, define
\begin{equation*}
	\mT_{K,\alpha}f(x)=\sum_{\tau\in\overline{\mI}_K}\frac{\partial^\alpha f(x_{\tau,K})}{\alpha!}\Psi_{\tau,K}(x)(x-x_{\tau,K})^\alpha.
\end{equation*}
We then have $\mT_{K,r}f=\sum_{\|\alpha\|_1=r}\mT_{K,\alpha}f$. Proposition~\ref{prop:fixed-high} gives the approximation of each such component. Taking the depth to be a sufficiently large constant multiple of $K^d\log_2(2K)$ makes its error at most $C_{d,s}K^{-s}$, while the width and norm product remain fixed.
\begin{proposition}[Task 4. Approximation of $\mT_{K,\alpha}f$]\label{prop:fixed-high}
	Fix $d\in\mathbb N^+$ and $s>d/2+1$, and set $u=\lceil s\rceil-1$. There exist constants $A_{d,s}\geq1$, $c_{d,s}>0$, $W_{d,s}\in\mathbb N^+$, and $\Gamma_{d,s}\geq1$ such that the following holds. Let $J\in\mathbb N^+$, $K=2^J$, and $f\in\mH^s([0,1]^d,1)$. For every $\alpha\in\mathbb N_0^d$ satisfying $d+1\leq r=\|\alpha\|_1\leq u$ and every integer $L\geq A_{d,s}K^d$, there exists a ReLU network $\Phi_{\alpha,J,L}\in\mF_{\text{\normalfont SSNN}}(L,W_{d,s},\Gamma_{d,s})$ such that
	\begin{equation*}
		\sup_{x\in[0,1]^d}\big|\Phi_{\alpha,J,L}(x)-\mT_{K,\alpha}f(x)\big|
		\leq2^{d+2}K^{-r}2^{-c_{d,s}L/K^d}.
	\end{equation*}
\end{proposition}

\section{Statistical estimation}\label{sec:estimation}
In this section, we introduce a penalized least squares estimator and establish non-asymptotic bounds on its prediction error. We then discuss applications to two-stage learning and adversarial robustness.

\subsection{Penalized least squares estimation}
\label{subsec:estimation}

We consider least squares estimators of $f_0$ based on $\mathcal D_n$. Let $(X,Y)$ be an independent copy of any observation in $\mathcal D_n$. For a given regression function $f$, define its population risk as $\mathcal R(f)= E_{(X,Y)}\{Y-f(X)\}^2$. Under model~\eqref{equ:model}, we have $\mathbb E(\varepsilon\mid X)=0$ and
\begin{equation*}
	\mathcal R(f)-\mathcal R(f_0) =	\mathbb E_X\big\{f(X)-f_0(X)\big\}^2.
\end{equation*}
Therefore, $f_0$ minimizes the population risk, and the excess risk equals the squared prediction error. The corresponding empirical risk is given by $\mathcal R_n(f) = \sum_{i=1}^n\{Y_i-f(X_i)\}^2/n$.  For a given function class $\mF_n$, an empirical risk minimizer is defined by $\wh{f}_n\in\argmin_{f\in\mF_n}\mathcal R_n(f)$. 

In this work, we take $\mF_n$ to be a class of feedforward neural networks. To be specific, for any integers $\mD\geq1$ and $\mW\geq d$, we use $\mF_n=\mF_{\text{FNN}}(\mD,\mW)$, and let $\Theta_{\mD,\mW}$ denote its collection of network parameters. For every $\Theta\in\Theta_{\mD,\mW}$, write $f_\Theta$ for the corresponding network function \eqref{equ:fnn}. Recall that the norm product $\rho(f_\Theta)$ defined in \eqref{equ:rhonorm} is computable from the network parameters and satisfies $\text{Lip}(f_\Theta)\leq\rho(f_\Theta)$. To preserve the Lipschitz property, we penalize the norm product. In the minimizations below, each network is understood together with its specified parameter representation, and $\rho(f)$ denotes the SSNP of that representation. For a penalty parameter $\lambda_n>0$, define the penalized least squares loss function as 
\begin{equation*}
	\mQ_n(f) = \mathcal R_n(f)+\lambda_n\rho(f).
\end{equation*}
The parameter $\lambda_n$ determines the strength of the penalty on the norm product. Then, the penalized least squares estimator is defined by
\begin{equation*}
	\wh{f}_n^{\text{P}}\in\argmin_{f\in\mF_n}\mQ_n(f).
\end{equation*}
Here we assume that a measurable minimizer exists. The network dimensions $\mD,\mW$ and the penalty parameter $\lambda_n$ may depend on $n$. We evaluate $\wh f_n^{\text{P}}$ by its expected excess risk $\mathbb E_{\mathcal D_n}\{\mathcal R(\wh f_n^{\text{P}})-\mathcal R(f_0)\}$. Our objective is to establish non-asymptotic upper bounds for this expected prediction error.

\subsection{Non-asymptotic error bounds}\label{subsec:error}

To study the prediction error of $\wh f_n^{\text{P}}$, we start from a basic decomposition. By the definition of $\wh f_n^{\text{P}}$, we know that $\mQ_n(\wh f_n^{\text{P}}) =\inf_{f\in\mF_n}\{\mathcal R_n(f) + \lambda_n\rho(f)\}$. Since $\mathbb E_{\mD_n}\{\mathcal R_n(f)\}=\mathcal R(f)$ for any fixed $f\in\mF_n$, we can obtain that
\begin{gather}\label{equ:est-risk-decomposition}
	\begin{aligned}
		&\mathbb E_{\mathcal D_n}\Big\{\mathcal R(\wh f_n^{\text{P}})-\mathcal R(f_0)\Big\}\\
		\leq \mathbb E_{\mathcal D_n}\Big\{\mathcal R(\wh f_n^{\text{P}}) -\mathcal R_n(\wh f_n^{\text{P}}) &-\lambda_n\rho(\wh f_n^{\text{P}})\Big\} + \inf_{f\in\mF_n} \Big\{\mathcal R(f)-\mathcal R(f_0) +\lambda_n\rho(f)\Big\}.
	\end{aligned}
\end{gather}
The first term is the stochastic contribution adjusted by the fitted penalty. The second term is the penalized approximation error over $\mF_n$. To bound the second term, we need a comparison network with both small approximation error and a controlled norm product. Therefore, we consider the subclass $\mF_{\text{SSNN}}(\mD,\mW,\mG)\subseteq\mF_n$ for some $\mG>0$. We then have $\inf_{f\in\mF_n} \big\{\mathcal R(f)-\mathcal R(f_0)+\lambda_n\rho(f)\big\}\leq \inf_{f\in\mF_{\text{SSNN}}(\mD,\mW,\mG)} \big\{\mathcal R(f)-\mathcal R(f_0)+\lambda_n\rho(f)\big\} \leq  \inf_{f\in\mF_{\text{SSNN}}(\mD,\mW,\mG)} \|f-f_0\|_\infty^2+\lambda_n\mG$. Theorem~\ref{thm:lnn-holder-approx} controls the approximation error on the right for suitable architectures. Note that the budget $\mG$ is used only for comparison. This leads to an upper bound on this term with suitable network architectures.

We next consider how to control the stochastic term in \eqref{equ:est-risk-decomposition}. Under model~\eqref{equ:model}, we know that $\mathbb E_{\mathcal D_n}\{\mathcal R_n(f_0)\}=\mathcal R(f_0)$. We then have
\begin{equation*}
	\mathbb E_{\mathcal D_n}\Big\{ \mathcal R(\wh f_n^{\text{P}}) -\mathcal R_n(\wh f_n^{\text{P}}) -\lambda_n\rho(\wh f_n^{\text{P}})\Big\} = \text{EP}_1 + \text{EP}_2,
\end{equation*}
where $\text{EP}_1 = \mathbb E_{\mathcal D_n}\big[ \mathcal R(\wh f_n^{\text{P}})-\mathcal R(f_0) -\sum_{i=1}^n\{\wh f_n^{\text{P}}(X_i)-f_0(X_i)\}^2/n \big]$ and $ \text{EP}_2 =  \mathbb E_{\mathcal D_n}\big[2\sum_{i=1}^n$ $\varepsilon_i \{\wh f_n^{\text{P}}(X_i)-f_0(X_i)\}/n -\lambda_n\rho(\wh f_n^{\text{P}}) \big]$. The first term $ \text{EP}_1 $ compares the population and empirical squared prediction errors. The second term $ \text{EP}_2$ contains the interaction between the noise and the fitted error after accounting for the penalty. Although $\mathbb E(\varepsilon_i\mid X_i)=0$, the  interaction term in $ \text{EP}_2$ need not have zero expectation because $\wh f_n^{\text{P}}$ depends on the same observations. To control both terms, we impose the following tail condition on the noise \citep{bauer2019deep,schmidt2020nonparametric}. This assumption allows unbounded responses. Since $f_0$ is bounded, it also implies that $\mathbb E\{\exp(c_Y Y^2)\}<\infty$ for some $c_Y>0$.
\begin{assumption}\label{ass:est-response-tail}
	There exists a constant $\nu_\varepsilon>0$ such that $\mathbb E\{\exp(u\varepsilon)\mid X\} \leq\exp(\nu_\varepsilon^2u^2/2)$ almost surely for every $u\in\mathbb R$.
\end{assumption}
%

Under Assumption~\ref{ass:est-response-tail}, it remains to bound $\mathrm{EP}_1+\mathrm{EP}_2$. Since $\wh{f}_n^\text{P}$ depends on the training sample, we require stochastic bounds that hold uniformly over all candidate networks. To address the penalty in $\mathrm{EP}_2$, we first derive these bounds over $\mF_{\text{SSNN}}(\mD,\mW,\mG)$ for each $\mG>0$. We quantify the complexity of these subclasses through covering numbers \citep{bartlett2017spectrally,neyshabur2018pac,golowich2020size,jiao2023deep}. For a function class $\mathcal F$ on $[0,1]^d$ and a positive number $\delta$, let $\mathcal N(\delta,\mathcal F,\|\cdot\|_\infty)$ denote the smallest number of uniform-norm balls of radius $\delta$ needed to cover $\mF$.  Since $\rho(f_\Theta)$ does not constrain the output bias, $ \mF_{\text{SSNN}}(\mD,\mW, \mG)$ contains all constant functions and has infinite covering numbers under the uniform norm. Therefore, we separate each $f_\Theta$ into the scalar shift $f_\Theta(0)$ and the centered part $f_\Theta-f_\Theta(0)$. The covering argument is applied to the centered class $ \mF_{\text{SSNN}}^0(\mD,\mW,\mG) = \big\{ f_\Theta-f_\Theta(0): f_\Theta\in\mF_{\text{SSNN}}(\mD,\mW,\mG) \big\}$. The scalar term $f_\Theta(0)$ is handled separately in the stochastic analysis. Lemma 14 in Appendix C gives
\begin{equation*}
		\log\mathcal N\Big( \delta,\mF_{\text{SSNN}}^0(\mD,\mW,\mG), \|\cdot\|_\infty \Big)
		\leq C_d\mD\mW^2 \log\left(1+\frac{\mG\mD\sqrt{\mW}}{\delta}\right)
\end{equation*}
for any $\delta>0$, where $C_d>0$ is a fixed constant. For fixed width and norm budget, this upper bound is linear in depth apart from a logarithmic factor. The resulting noise bounds are combined across norm-product budgets and used with penalized optimality to control the fitted norm product. This control then permits an empirical-to-population comparison based on the same entropy bound. Combining these stochastic bounds with Theorem~\ref{thm:lnn-holder-approx} leads to the following theorem. Theorem~\ref{thm:lnn-estimation} bounds the expected excess risk and the expected fitted norm product for a specified architecture with a suitable penalty level. 
\begin{theorem}\label{thm:lnn-estimation}
	Under model~\eqref{equ:model}, suppose that $f_0\in\mH^s([0,1]^d,B_0)$ with $s>d/2+1$ and $B_0>0$. With Assumption~\ref{ass:est-response-tail}, there exist an integer $W_*\geq d$ and positive constants $c_D,c_\lambda,C_R,C_\rho$, such that the following holds. Set $D_n=\left\lceil\{n/\{\log(en)\}^2\}^{d/(2s+d)}\right\rceil$. Choose the network depth, width and penalty parameter as $\mD=\left\lceil c_DD_n\log_2(2D_n)\right\rceil $, $ \mW=W_*$, and $ \lambda_n=c_\lambda n^{-2s/(2s+d)}\{\log(en)\}^{4s/(2s+d)}$, respectively. We then have
	\begin{equation}\label{equ:est-expected-risk}
		\mathbb E_{\mathcal D_n}\big\{\mathcal R(\wh f_n^{\text{P}})-\mathcal R(f_0)\big\} \leq C_R n^{-2s/(2s+d)}\big\{\log(en)\big\}^{4s/(2s+d)}
	\end{equation}
	and $\mathbb E_{\mathcal D_n}\{\rho(\wh f_n^{\mathrm P})\}\leq C_\rho$.
\end{theorem}
The proof of Theorem~\ref{thm:lnn-estimation} is given in Appendix B.1. For the H\"older classes considered here, the theorem matches the optimal nonparametric prediction rate $n^{-2s/(2s+d)}$ up to a logarithmic factor \citep{schmidt2020nonparametric,jiao2023deep,bhattacharya2024deep}. This rate is attained with fixed width and depth of order $n^{d/(2s+d)}\{\log(en)\}^{(2s-d)/(2s+d)}$. The same choice of $\lambda_n$ also gives $\mathbb E_{\mathcal D_n}\{\rho(\wh f_n^{\text{P}})\}\leq C_\rho$. Thus, the expected norm product remains bounded as the prescribed network depth increases, even though $\lambda_n$ tends to zero. This bound is obtained through penalized estimation without imposing a fixed norm-product budget on the fitting class $\mF_n$. Since $\operatorname{Lip}(\wh f_n^{\text{P}})\leq\rho(\wh f_n^{\text{P}})$, it also implies $\mathbb E_{\mathcal D_n}\{\operatorname{Lip}(\wh f_n^{\text{P}})\}\leq C_\rho$. Therefore, the estimator obtains the stated prediction rate with an expected Lipschitz constant bounded.

\subsection{Applications}\label{subsec:applications}

Next, we present two applications of the preceding results. The first is two-stage learning with estimated covariates. The second is adversarial robustness to bounded perturbations of the test inputs. Throughout this subsection, write $r_n=n^{-2s/(2s+d)}\{\log(en)\}^{4s/(2s+d)}$ for the prediction rate in Theorem~\ref{thm:lnn-estimation}.

\textit{Two-stage learning.}
Statistical analyses of two-stage learning include regression with generated covariates \citep{mammen2012nonparametric}, factor-augmented neural networks \citep{fan2024factor}, and kernel distribution regression \citep{bachoc2025improved}. Here we study how first-stage covariate estimation errors propagate into the prediction error of the second-stage network. Let $X_i\in[0,1]^d$ denote the covariates in model~\eqref{equ:model} and let $\widehat X_i\in[0,1]^d$ be their first-stage estimates. Write $\mathcal A_N$ for the first-stage training information based on $N$ observations. Conditional on $\mathcal A_N$, assume that $(X_i,\widehat X_i,\varepsilon_i)$ with $i=1,\ldots,n$ are independent and identically distributed and that the test triple $(X,\widehat X,\varepsilon)$ is an independent observation from the same conditional distribution. The second stage fits a predictor using $(\widehat X_i,Y_i)$. For a new observation, the fitted predictor is evaluated at an estimated covariate $\widehat X$.  For any $f\in\mF_n$, we have $|f(\widehat X_i)-f(X_i)|\leq\rho(f)\|\widehat X_i-X_i\|_2$. Thus, the norm product controls the effect of covariate estimation error on the network output. For a prescribed architecture, define the two-stage penalized least squares estimator by
\begin{equation*}
	\wh f_n^{\text{TP}}
	\in\argmin_{f\in\mF_n}
	\left[
	\frac1n\sum_{i=1}^n\big\{Y_i-f(\widehat X_i)\big\}^2
	+\lambda_n\rho(f)
	\right].
\end{equation*}
The following corollary bounds its prediction error in terms of $r_n$ and the first-stage mean squared error. The proof of Corollary~\ref{coro:two-stage-regression} is given in Appendix B.2. The first-stage estimation error contributes the term $C\delta_N^2$ to the prediction bound. Its coefficient does not grow with the prescribed network depth. If $\delta_N^2=O(r_n)$, the estimator achieves the same prediction rate as when the covariates are observed.
\begin{corollary}[Two-stage learning]\label{coro:two-stage-regression}
Under the preceding sampling conditions, assume that $f_0\in\mH^s([0,1]^d,B_0)$ with $s>d/2+1$. Suppose that $\mathbb E\{\|\widehat X-X\|_2^2\mid\mathcal A_N\}\leq\delta_N^2$ almost surely for a deterministic $\delta_N\geq0$, and $\mathbb E\{\exp(u\varepsilon)\mid X,\widehat X,\mathcal A_N\}\leq\exp(\nu_\varepsilon^2u^2/2)$ almost surely for every $u\in\mR$, where $\nu_\varepsilon>0$ is fixed. Use the architecture in Theorem~\ref{thm:lnn-estimation}. There exist constants $c_\lambda,C>0$ such that choosing $\lambda_n=c_\lambda(r_n+\delta_N^2)$ gives, for every $n>1$,
	\begin{equation*}
		\mathbb E\Big[
		\{\wh f_n^{\text{TP}}(\widehat X)-f_0(X)\}^2
		\mid\mathcal A_N
		\Big]
		\leq C(r_n+\delta_N^2)
	\end{equation*}
	almost surely. The expectation integrates over the second-stage training sample and the independent test pair, conditional on $\mathcal A_N$.
\end{corollary}

\textit{Adversarial robustness.}
Statistical analyses of adversarial prediction include linear regression \citep{javanmard2020precise} and nonparametric regression under a supremum-norm loss \citep{moradi2026adversarial}. Here we study prediction under bounded perturbations of the test inputs. Consider the estimator $\wh f_n^{\text{P}}$ fitted to the unperturbed sample $\mathcal D_n$. At prediction time, an input $X$ may be replaced by any $Z\in[0,1]^d$ satisfying $\|Z-X\|_2\leq\eta$, where $\eta\geq0$ is the perturbation radius. The target remains $f_0(X)$. For any $f\in\mF_n$, define its worst-case squared prediction error by
\begin{equation*}
	\mathcal E_\eta(f)
	=
	\mathbb E_X\left[
	\sup_{Z\in[0,1]^d, \|Z-X\|_2\leq\eta}
	\big\{f(Z)-f_0(X)\big\}^2
	\right].
\end{equation*}
At $\eta=0$, this reduces to the usual prediction error $\mathcal E_0(f)=\mathcal R(f)-\mathcal R(f_0)$. The supremum allows the perturbation to depend on both $X$ and the network $f$. Note that no distributional assumption on the perturbation is needed. The bound $|f(Z)-f(X)|\leq\eta\rho(f)$ gives
\begin{equation*}
	\mathcal E_\eta(f)
	\leq
	2\{\mathcal R(f)-\mathcal R(f_0)\}
	+2\eta^2\rho(f)^2.
\end{equation*}
Thus, the norm product controls the perturbation term in the prediction bound. Corollary~\ref{coro:adversarial-robustness} gives the corresponding bound on the expected worst-case prediction error of $\wh f_n^{\text{P}}$. The proof is given in the Appendix B.3. When $\eta^2\leq r_n$, the bound gives the same prediction rate as in the unperturbed setting. The coefficient of $\eta^2$ does not grow as the network depth increases. This guarantee holds for the original penalized estimator without adversarial training.
\begin{corollary}[Adversarial robustness]\label{coro:adversarial-robustness}
	Suppose that the assumptions of Theorem~\ref{thm:lnn-estimation} hold, and use the architecture and penalty level specified there. There exists a constant $C>0$ such that
	\begin{equation*}
		\mathbb E_{\mathcal D_n}
		\big\{\mathcal E_\eta(\wh f_n^{\text{P}})\big\}
		\leq C(r_n+\eta^2),
	\end{equation*}
	for every $\eta\geq0$.
\end{corollary}

\section{Numerical studies}\label{sec:simulation}

To demonstrate the finite-sample performance of the proposed estimator, we conduct a number of simulation studies. We evaluate its estimation accuracy and sensitivity to input perturbations, and examine how estimation error changes with sample size. 

\subsection{The simulation setup}\label{subsec:simulation-settings}

We generate $n$ independent observations $\{X_i,Y_i\}_{i=1}^n$ from model~\eqref{equ:model} (i.e., $Y_i=f_0(X_i)+\varepsilon_i$). Here the covariate vector $X_i=(X_{i1},\ldots,X_{id})^\top$ has independent components $X_{ik}$ drawn from the uniform distribution on $[0,1]$ and the errors $\varepsilon_i$ are independently sampled from $N(0,\sigma_\varepsilon^2)$. To cover regression functions with different structures, we consider the following three choices of $f_0$
\begin{equation*}
	\begin{aligned}
		f_{\mathrm{A},d}(x)&=-\frac{1}{2d}\sum_{j=1}^d u_j^{(3)}(x),\\
		f_{\mathrm{B},d}(x)&=\frac12\prod_{j=1}^d\cos(4\pi x_j),\\
		f_{\mathrm{C},d}(x)&=\frac{0.3}{\sqrt d}\sum_{j=1}^d\{\sin(2\pi x_j)+0.3\sin(6\pi x_j)\},
	\end{aligned}
\end{equation*}
where $u_j^{(0)}(x)=\cos(\pi x_j)$ and $u_j^{(r+1)}(x)=2\{0.6u_j^{(r)}(x)+0.4\sum_{k=1}^d u_k^{(r)}(x)/d\}^{2}-1$ for $ r=0,1,2$. We refer to the resulting models as DGP~A, DGP~B and DGP~C, respectively. Here DGP A has a recursive compositional structure \citep{telgarsky2016benefits}, DGP B has a multiplicative structure \citep{yan2025deep}, and DGP C has an additive structure \citep{dai2023kernel}. For each DGP, we consider the sample sizes $n\in\{500,1000,2000\}$, the covariate dimensions $d\in\{1,2\}$ and the noise levels $\sigma_\varepsilon\in\{0.1,0.2,0.3,0.4\}$.

For each simulation setting, the penalized estimator $\wh f_n^{\mathrm{P}}$ introduced in Section~\ref{subsec:estimation} is fitted over fully connected ReLU networks with width $\mW=8$ and hidden depth $\mD\in\{2,4,6,8,10\}$, where the network class with depth $\mD$ is denoted by $\mF_{\text{FNN}}(\mD,8)$. The penalty parameter $\lambda$ takes values in $\{0,3\times10^{-6},2\times10^{-5},5\times10^{-5},7\times10^{-5},10^{-4},3\times10^{-4},10^{-3},3\times10^{-3}\}$, specified on the original response scale. To assess the benefit of the penalty, we compare $\wh f_n^{\text{P}}$ with the unpenalized least squares estimator $\wh f_n\in\argmin_{f\in\mF_{\text{FNN}}(\mD,8)}\sum_{i=1}^n\{Y_i-f(X_i)\}^2/n$, which is fitted over the same network class so that the comparison is not affected by network capacity. In each replication, we use $80\%$ of the observations for training and the remaining $20\%$ for validation. Both estimators are fitted using the training observations and the Adam optimizer. The depth of $\wh f_n$ is selected by minimizing the validation mean squared error, and the depth and the penalty parameter of $\wh f_n^{\text{P}}$ are selected jointly on the validation set.

\subsection{Simulation results}\label{subsec:simulation-results}

We first compare the penalized estimator $\wh f_n^{\text{P}}$ (Penalized) with the unpenalized estimator $\wh f_n$ (FNN) in terms of estimation accuracy and input sensitivity. Both estimators are fitted over the same network class. First, for an estimator $\wh f\in\{\wh f_n,\wh f_n^{\mathrm{P}}\}$, estimation accuracy is measured by the test error
\begin{equation*}
	\operatorname{TE}(\wh f) = \frac{1}{N_{\text{test}}}	\sum_{j=1}^{N_{\text{test}}} \big\{\wh f(X_j^{\text{test}}) -f_0(X_j^{\text{test}})\big\}^{2},
\end{equation*}
where the test inputs $X_j^{\text{test}}$ are sampled independently from uniform distribution on $[0,1]^d$ and $N_{\text{test}}=10000$. They are independent of the training and validation data. Within each replication, both estimators are evaluated on the same test inputs. A smaller test error indicates more accurate estimation. Second, input sensitivity is measured by
\begin{equation*}
	S_{\max}(\wh f) = \max_{x\in\mathcal G_d}\max_{u\in\mathcal U_d} \frac{|\wh f(x+\epsilon u)-\wh f(x)|}{\epsilon},
\end{equation*}
which is the largest change of the fitted values per unit input perturbation over a set of grid points and perturbation directions. Here $\epsilon=0.005$, and $\mathcal G_d$ is an equally spaced grid on $[0.01,0.99]^d$ with $2001$ points along each coordinate, which keeps all perturbed inputs inside $[0,1]^d$. The perturbation directions are given by $\mathcal U_d=\{v/\|v\|_2:v\in\{-1,0,1\}^d\setminus\{\boldsymbol0\}\}$. A smaller value of $S_{\max}$ indicates lower sensitivity, and we also compute $S_{\max}(f_0)$ as a reference level for the true regression function.

For each DGP and each choice of $(n,d,\sigma_\varepsilon)$, we repeat the experiment $R=100$ times. For each estimator, let $\operatorname{TE}^{(r)}$ and $S_{\max}^{(r)}$ denote the test error and input sensitivity obtained in the $r$th replication with $1\leq r\leq R$, respectively. We define the mean test error and mean input sensitivity as $\text{MTE} = \sum_{r=1}^{R}\operatorname{TE}^{(r)}/R$ and $ \text{MS}_{\max} = \sum_{r=1}^{R}S_{\max}^{(r)}/R$, respectively. Table~\ref{tab:simulation-current-core} reports these means for $n=500$, with Monte Carlo standard errors (SEs) in parentheses. It includes all three DGPs, both dimensions and all four noise levels. The column labelled True gives $S_{\max}(f_0)$. Results for $n=1000$ and $2000$ are reported in Table S.1 of the supplementary material. 
\begin{table}[t]
	\centering
	\caption{Estimation error and input sensitivity at $n=500$. Bold marks the smaller result.}\label{tab:simulation-current-core}
	\begingroup
	\small
	\setlength{\tabcolsep}{3pt}
	\renewcommand{\arraystretch}{1.12}
	\begin{tabular}{@{}crr rrrrr@{}}
		\toprule
		& & & \multicolumn{2}{c}{$\text{MTE\,(SE)}\,(10^{-3})$} & \multicolumn{3}{c}{$\text{MS}_{\max}\,$(SE)} \\
		\cmidrule(lr){4-5}\cmidrule(l){6-8}
		Function & $d$ & $\sigma_\varepsilon$ & \multicolumn{1}{c}{FNN} & \multicolumn{1}{c}{Penalized} & \multicolumn{1}{c}{FNN} & \multicolumn{1}{c}{{Penalized}} & \multicolumn{1}{c}{True} \\
		\midrule
		\multirow{8}{*}[-1.5pt]{$f_{\mathrm{A},d}$} & 1 & 0.1 & \makebox[3.0em][r]{\textbf{1.08}}\,\makebox[3.3em][r]{(0.02)} & \makebox[3.0em][r]{1.13}\,\makebox[3.3em][r]{(0.04)} & \makebox[3.0em][r]{20.75}\,\makebox[3.3em][r]{(0.60)} & \makebox[3.0em][r]{\textbf{14.27}}\,\makebox[3.3em][r]{(0.20)} & 12.56 \\
		& 1 & 0.2 & \makebox[3.0em][r]{\textbf{3.28}}\,\makebox[3.3em][r]{(0.10)} & \makebox[3.0em][r]{3.64}\,\makebox[3.3em][r]{(0.17)} & \makebox[3.0em][r]{23.52}\,\makebox[3.3em][r]{(0.94)} & \makebox[3.0em][r]{\textbf{14.24}}\,\makebox[3.3em][r]{(0.32)} & 12.56 \\
		& 1 & 0.3 & \makebox[3.0em][r]{\textbf{6.45}}\,\makebox[3.3em][r]{(0.18)} & \makebox[3.0em][r]{6.92}\,\makebox[3.3em][r]{(0.31)} & \makebox[3.0em][r]{28.45}\,\makebox[3.3em][r]{(1.37)} & \makebox[3.0em][r]{\textbf{14.46}}\,\makebox[3.3em][r]{(0.34)} & 12.56 \\
		& 1 & 0.4 & \makebox[3.0em][r]{\textbf{11.00}}\,\makebox[3.3em][r]{(0.36)} & \makebox[3.0em][r]{11.25}\,\makebox[3.3em][r]{(0.62)} & \makebox[3.0em][r]{34.34}\,\makebox[3.3em][r]{(1.98)} & \makebox[3.0em][r]{\textbf{15.18}}\,\makebox[3.3em][r]{(0.53)} & 12.56 \\
		\addlinespace[3pt]
		& 2 & 0.1 & \makebox[3.0em][r]{8.00}\,\makebox[3.3em][r]{(0.23)} & \makebox[3.0em][r]{\textbf{7.78}}\,\makebox[3.3em][r]{(0.18)} & \makebox[3.0em][r]{24.25}\,\makebox[3.3em][r]{(1.09)} & \makebox[3.0em][r]{\textbf{12.44}}\,\makebox[3.3em][r]{(0.43)} & 8.88 \\
		& 2 & 0.2 & \makebox[3.0em][r]{18.61}\,\makebox[3.3em][r]{(0.46)} & \makebox[3.0em][r]{\textbf{16.08}}\,\makebox[3.3em][r]{(0.39)} & \makebox[3.0em][r]{31.20}\,\makebox[3.3em][r]{(2.50)} & \makebox[3.0em][r]{\textbf{9.66}}\,\makebox[3.3em][r]{(0.43)} & 8.88 \\
		& 2 & 0.3 & \makebox[3.0em][r]{29.49}\,\makebox[3.3em][r]{(0.72)} & \makebox[3.0em][r]{\textbf{24.50}}\,\makebox[3.3em][r]{(0.65)} & \makebox[3.0em][r]{25.30}\,\makebox[3.3em][r]{(2.52)} & \makebox[3.0em][r]{\textbf{9.36}}\,\makebox[3.3em][r]{(0.54)} & 8.88 \\
		& 2 & 0.4 & \makebox[3.0em][r]{41.21}\,\makebox[3.3em][r]{(1.12)} & \makebox[3.0em][r]{\textbf{34.35}}\,\makebox[3.3em][r]{(0.80)} & \makebox[3.0em][r]{25.04}\,\makebox[3.3em][r]{(2.85)} & \makebox[3.0em][r]{\textbf{7.26}}\,\makebox[3.3em][r]{(0.61)} & 8.88 \\
		\midrule
		\multirow{8}{*}[-1.5pt]{$f_{\mathrm{B},d}$} & 1 & 0.1 & \makebox[3.0em][r]{\textbf{0.68}}\,\makebox[3.3em][r]{(0.03)} & \makebox[3.0em][r]{0.72}\,\makebox[3.3em][r]{(0.04)} & \makebox[3.0em][r]{10.02}\,\makebox[3.3em][r]{(0.55)} & \makebox[3.0em][r]{\textbf{6.63}}\,\makebox[3.3em][r]{(0.15)} & 6.28 \\
		& 1 & 0.2 & \makebox[3.0em][r]{\textbf{1.93}}\,\makebox[3.3em][r]{(0.07)} & \makebox[3.0em][r]{2.24}\,\makebox[3.3em][r]{(0.12)} & \makebox[3.0em][r]{10.77}\,\makebox[3.3em][r]{(0.67)} & \makebox[3.0em][r]{\textbf{6.31}}\,\makebox[3.3em][r]{(0.23)} & 6.28 \\
		& 1 & 0.3 & \makebox[3.0em][r]{\textbf{3.78}}\,\makebox[3.3em][r]{(0.19)} & \makebox[3.0em][r]{3.80}\,\makebox[3.3em][r]{(0.16)} & \makebox[3.0em][r]{11.87}\,\makebox[3.3em][r]{(1.04)} & \makebox[3.0em][r]{\textbf{6.07}}\,\makebox[3.3em][r]{(0.14)} & 6.28 \\
		& 1 & 0.4 & \makebox[3.0em][r]{\textbf{5.73}}\,\makebox[3.3em][r]{(0.31)} & \makebox[3.0em][r]{5.79}\,\makebox[3.3em][r]{(0.28)} & \makebox[3.0em][r]{14.18}\,\makebox[3.3em][r]{(2.12)} & \makebox[3.0em][r]{\textbf{6.10}}\,\makebox[3.3em][r]{(0.17)} & 6.28 \\
		\addlinespace[3pt]
		& 2 & 0.1 & \makebox[3.0em][r]{14.13}\,\makebox[3.3em][r]{(0.36)} & \makebox[3.0em][r]{\textbf{12.59}}\,\makebox[3.3em][r]{(0.27)} & \makebox[3.0em][r]{32.96}\,\makebox[3.3em][r]{(1.71)} & \makebox[3.0em][r]{\textbf{15.14}}\,\makebox[3.3em][r]{(0.43)} & 6.28 \\
		& 2 & 0.2 & \makebox[3.0em][r]{26.49}\,\makebox[3.3em][r]{(0.59)} & \makebox[3.0em][r]{\textbf{23.16}}\,\makebox[3.3em][r]{(0.46)} & \makebox[3.0em][r]{34.89}\,\makebox[3.3em][r]{(2.22)} & \makebox[3.0em][r]{\textbf{13.65}}\,\makebox[3.3em][r]{(0.60)} & 6.28 \\
		& 2 & 0.3 & \makebox[3.0em][r]{39.77}\,\makebox[3.3em][r]{(0.89)} & \makebox[3.0em][r]{\textbf{34.39}}\,\makebox[3.3em][r]{(0.72)} & \makebox[3.0em][r]{33.94}\,\makebox[3.3em][r]{(2.60)} & \makebox[3.0em][r]{\textbf{12.27}}\,\makebox[3.3em][r]{(0.76)} & 6.28 \\
		& 2 & 0.4 & \makebox[3.0em][r]{53.96}\,\makebox[3.3em][r]{(1.32)} & \makebox[3.0em][r]{\textbf{46.12}}\,\makebox[3.3em][r]{(0.90)} & \makebox[3.0em][r]{31.70}\,\makebox[3.3em][r]{(3.14)} & \makebox[3.0em][r]{\textbf{8.05}}\,\makebox[3.3em][r]{(0.57)} & 6.28 \\
		\midrule
		\multirow{8}{*}[-1.5pt]{$f_{\mathrm{C},d}$} & 1 & 0.1 & \makebox[3.0em][r]{\textbf{0.64}}\,\makebox[3.3em][r]{(0.03)} & \makebox[3.0em][r]{0.82}\,\makebox[3.3em][r]{(0.06)} & \makebox[3.0em][r]{9.15}\,\makebox[3.3em][r]{(0.91)} & \makebox[3.0em][r]{\textbf{4.14}}\,\makebox[3.3em][r]{(0.33)} & 3.58 \\
		& 1 & 0.2 & \makebox[3.0em][r]{\textbf{1.92}}\,\makebox[3.3em][r]{(0.07)} & \makebox[3.0em][r]{2.26}\,\makebox[3.3em][r]{(0.11)} & \makebox[3.0em][r]{10.07}\,\makebox[3.3em][r]{(1.10)} & \makebox[3.0em][r]{\textbf{4.10}}\,\makebox[3.3em][r]{(0.56)} & 3.58 \\
		& 1 & 0.3 & \makebox[3.0em][r]{\textbf{3.40}}\,\makebox[3.3em][r]{(0.17)} & \makebox[3.0em][r]{3.40}\,\makebox[3.3em][r]{(0.14)} & \makebox[3.0em][r]{12.59}\,\makebox[3.3em][r]{(1.71)} & \makebox[3.0em][r]{\textbf{3.61}}\,\makebox[3.3em][r]{(0.58)} & 3.58 \\
		& 1 & 0.4 & \makebox[3.0em][r]{5.50}\,\makebox[3.3em][r]{(0.34)} & \makebox[3.0em][r]{\textbf{4.53}}\,\makebox[3.3em][r]{(0.20)} & \makebox[3.0em][r]{13.89}\,\makebox[3.3em][r]{(2.16)} & \makebox[3.0em][r]{\textbf{3.61}}\,\makebox[3.3em][r]{(0.57)} & 3.58 \\
		\addlinespace[3pt]
		& 2 & 0.1 & \makebox[3.0em][r]{2.88}\,\makebox[3.3em][r]{(0.06)} & \makebox[3.0em][r]{\textbf{2.85}}\,\makebox[3.3em][r]{(0.07)} & \makebox[3.0em][r]{7.28}\,\makebox[3.3em][r]{(0.52)} & \makebox[3.0em][r]{\textbf{3.45}}\,\makebox[3.3em][r]{(0.18)} & 3.58 \\
		& 2 & 0.2 & \makebox[3.0em][r]{5.46}\,\makebox[3.3em][r]{(0.18)} & \makebox[3.0em][r]{\textbf{5.13}}\,\makebox[3.3em][r]{(0.14)} & \makebox[3.0em][r]{6.58}\,\makebox[3.3em][r]{(0.88)} & \makebox[3.0em][r]{\textbf{2.62}}\,\makebox[3.3em][r]{(0.17)} & 3.58 \\
		& 2 & 0.3 & \makebox[3.0em][r]{7.79}\,\makebox[3.3em][r]{(0.19)} & \makebox[3.0em][r]{\textbf{7.16}}\,\makebox[3.3em][r]{(0.18)} & \makebox[3.0em][r]{4.50}\,\makebox[3.3em][r]{(0.38)} & \makebox[3.0em][r]{\textbf{2.32}}\,\makebox[3.3em][r]{(0.16)} & 3.58 \\
		& 2 & 0.4 & \makebox[3.0em][r]{12.43}\,\makebox[3.3em][r]{(0.80)} & \makebox[3.0em][r]{\textbf{9.14}}\,\makebox[3.3em][r]{(0.26)} & \makebox[3.0em][r]{6.58}\,\makebox[3.3em][r]{(1.23)} & \makebox[3.0em][r]{\textbf{2.36}}\,\makebox[3.3em][r]{(0.17)} & 3.58 \\
		\bottomrule
	\end{tabular}
	\endgroup
\end{table}

By Table~\ref{tab:simulation-current-core}, we obtain the following two findings. First, the penalized estimator substantially reduces input sensitivity. Its $\text{MS}_{\max}$ is smaller than that of FNN in all $24$ settings, with reductions ranging from about $31\%$ to $75\%$. Its $\text{MS}_{\max}$ is also closer to $S_{\max}(f_0)$ than that of FNN in $23$ of the $24$ settings. Second, this reduction in input sensitivity is achieved with comparable estimation accuracy. For example, in DGP A with $d=1$ and $\sigma_\varepsilon=0.4$, MTE increases from $11.00\times10^{-3}$ for FNN to $11.25\times10^{-3}$ for the penalized estimator, whereas $\text{MS}_{\max}$ decreases from $34.34$ to $15.18$. Thus, an increase of about $2\%$ in MTE is accompanied by a reduction of about $56\%$ in input sensitivity. These results indicate that the penalty provides better control of input sensitivity with estimation accuracy comparable to FNN.

\begin{figure}[tb]
	\centering
	\includegraphics[width=5.5in]{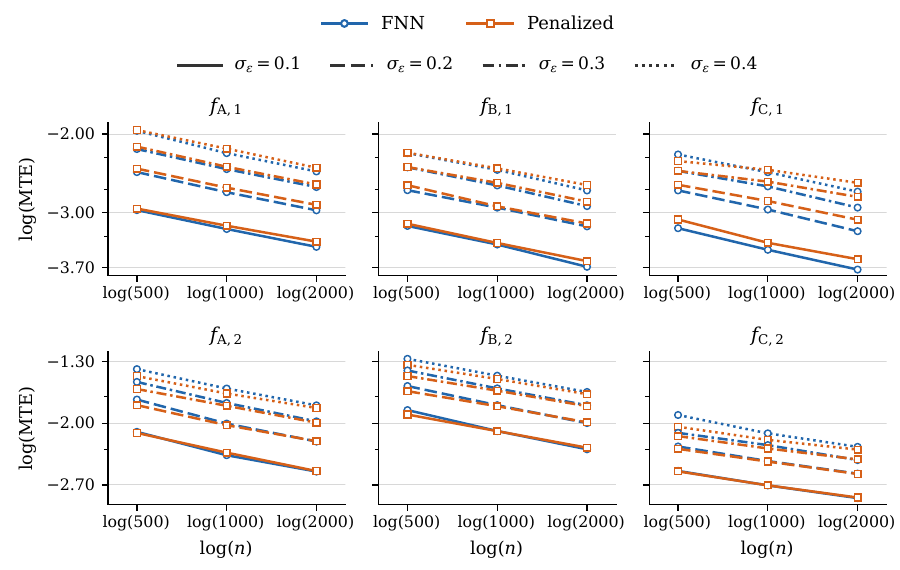}
	\caption{MTE of FNN and penalized estimator. The three columns correspond to $f_{\mathrm{A},d}$ (left), $f_{\mathrm{B},d}$ (middle) and $f_{\mathrm{C},d}$ (right). The top and bottom rows correspond to $d=1$ and $d=2$, respectively. Within each panel, blue circles represent FNN and orange squares represent penalized estimator. Different line styles indicate different noise levels.}
	\label{fig:simulation-current-convergence}
\end{figure}
Next, we consider how the estimation errors of the two estimators decrease as the sample size increases. Figure~\ref{fig:simulation-current-convergence} reports MTE for $n\in\{500,1000,2000\}$ under each DGP, dimension and noise level. Both axes show base-10 logarithms, so the curve slopes reflect empirical rates of error decay. MTE decreases with sample size for both estimators in all settings. The corresponding curves have similar slopes in most cases, suggesting comparable rates over the sample sizes considered.

\section{Conclusion and Discussion}\label{sec:con}

This paper develops approximation and estimation theory for deep ReLU networks with Lipschitz control. We introduce a split spectral norm product that provides a computable upper bound on the Lipschitz constant. For sufficiently smooth H\"{o}lder functions, our construction improves uniform approximation accuracy as depth increases, with both network width and the norm product bounded by fixed constants. Based on this result, we propose a penalized least squares estimator that achieves the optimal prediction rate $n^{-2s/(2s+d)}$, up to a logarithmic factor. Its expected Lipschitz constant is bounded uniformly in the sample size. We also establish prediction bounds for two-stage learning and adversarial input perturbations. In these bounds, the additional error terms have coefficients that do not grow with network depth. Numerical results show that the penalty reduces input sensitivity with prediction accuracy comparable to that of unpenalized networks.

To conclude this article, we wish to discuss several interesting topics for future research. First, our results focus on sufficiently smooth H\"{o}lder functions. Extending the approximation and estimation theory to broader classes of smooth functions is worth pursuing. Second, the theoretical choices of network depth and penalty level depend on the smoothness of the regression function. How to select these parameters by data-driven methods would be another interesting topic for future study. Finally, it would also be useful to extend our analysis to high-dimensional settings. This would help clarify how dimension affects prediction accuracy and input sensitivity.

\begin{supplement}
\stitle{Supplement to ``Optimal Deep Nonparametric Regression with Uniformly Bounded Lipschitz Constants'' \normalfont{(DOI: .pdf).} The supplementary material contains proofs of the main results, supporting lemmas and their proofs, and additional numerical results.}
\end{supplement} 

%

\bibliographystyle{imsart-nameyear} 
\bibliography{bibliography}       

\end{document}